\documentclass[twoside]{article}

\usepackage{PRIMEarxiv}

\usepackage[utf8]{inputenc} 
\usepackage[T1]{fontenc}    
\usepackage[colorlinks=true]{hyperref}       
\usepackage{url}            
\usepackage{booktabs}       
\usepackage{amsfonts}       
\usepackage{nicefrac}       
\usepackage{microtype}      
\usepackage{lipsum}
\usepackage{fancyhdr}       
\usepackage{graphicx}       
\graphicspath{{media/}}     
\usepackage{amsthm}
\usepackage{algorithm}
\usepackage{algorithmic}
\usepackage{amsmath}

\usepackage[style=ieee-alphabetic,sorting=nyvt]{biblatex}
\title{\texttt{BoxRL-NNV}: Boxed Refinement of Latin Hypercube Samples for Neural Network Verification}

\author{
  Sarthak Das \\
  Junior Research Fellow \\
  Indian Institute of Technology Gandhinagar, India\\
  \texttt{sarthak.das@iitgn.ac.in} \\
}

\begin{document}
\maketitle

\begin{abstract}
\texttt{BoxRL-NNV} is a Python tool for the detection of safety violations in neural networks by computing the bounds of the output variables, given the bounds of the input variables of the network. This is done using global extrema estimation via Latin Hypercube Sampling, and further refinement using L-BFGS-B for local optimization around the initial guess. This paper presents an overview of \texttt{BoxRL-NNV}, as well as our results for a subset of the ACAS Xu benchmark. A complete evaluation of the tool's performance, including benchmark comparisons with state-of-the-art tools, shall be presented at the Sixth International Verification of Neural Networks Competition (VNN-COMP’25).
\end{abstract}


\section{Introduction}

\texttt{BoxRL-NNV} (available at \href{https://github.com/dassarthak18/R2X-NNV/tree/BoxRL-NNV}{\texttt{https://github.com/dassarthak18/R2X-NNV/tree/BoxRL-NNV}}) is a tool written in Python for the detection of safety violations in neural networks. The software takes as inputs a neural network given in an ONNX (Open Neural Network Exchange) format, and a safety specification given as a VNNLIB file. Thereafter, \texttt{BoxRL-NNV} verifies whether the given neural network satisfies the safety properties specified by the VNNLIB file.

ONNX is an industry standard format for interchange of neural networks between different frameworks such as PyTorch and Tensorflow. VNNLIB \cite{FoMLAS2023:Supporting_Standardization_Neural_Networks}, likewise, is an international benchmarks standard for the verification of neural networks, which specifies safety properties as propositional logic satisfiability problems written in a subset of the SMT-LIB2 standard \cite{BarFT-RR-17}.

\section{Related Work}
Neural networks, while powerful, are inherently complex structures that can be regarded as multi-input multi-output (MIMO) black boxes. As such, interpreting them becomes very difficult. With their heavy deployment in a wide variety of safety-critical domains such as healthcare and autonomous navigation, it is becoming increasingly necessary to build trust and accountability in their use \cite{zhang2021survey}.

One approach is to leverage surrogate models such as decision trees \cite{yang2018deepneuraldecisiontrees} and Gaussian processes \cite{pmlr-v216-li23c} to increase interpretability on a global level, or use sophisticated model-agnostic methods such as LIME \cite{ribeiro2016should} and SHAP \cite{lundberg2017unifiedapproachinterpretingmodel} to generate local explanations for a given prediction. Another promising approach is neural network verification, which generates mathematical guarantees that a neural network respects its safety specifications, such as input-output bounds.

With the advent of friendly competitions such as International Verification of Neural Networks Competition (VNN-COMP) \cite{brix2023fourthinternationalverificationneural}, the problem of safety verification of neural networks is becoming more standardized, and we are seeing a shift from theoretical approaches to practical, measurable efforts. This tool, much like current state-of-the-art such as Marabou \cite{wu2024marabou}, $\alpha,\beta$-crown \cite{abcrown} and NeuralSAT \cite{duong2023dpll}, is a step in this direction.

\section{Methodology}

\texttt{BoxRL-NNV} treats neural networks as non-convex multi-input multi-output (MIMO) black boxes. As such, its verification algorithm assumes limited resources (no GPU acceleration) and no domain-specific knowledge (no encoding of the neural network architecture).

It extracts input bounds for any given neural network directly from the VNNLIB file and generates a sample of input points using Latin Hypercube Sampling (LHS), which is a Monte Carlo simulation method used to generate a near-random sample of parameter values from a multidimensional distribution \cite{ef76b040-2f28-37ba-b0c4-02ed99573416}. LHS is scalable and requires fewer samples to achieve the same level of accuracy as uniform sampling. This makes it particularly useful in complex simulations where computational resources are limited. Moreover, LHS ensures that samples are more evenly distributed across the range of each variable, reducing the correlation between samples and ensuring a better coverage of the entire distribution. In particular, we use a variant called LHS with multidimensional uniformity (LHSMDU) \cite{DEUTSCH2012763} for better coverage, and therefore stronger guarantees.

By computing the neural network outputs across these points, \texttt{BoxRL-NNV} identifies promising regions where global optima might be found. Thereafter, \texttt{BoxRL-NNV} picks the most promising region and performs a limited-memory boxed BFGS (L-BFGS-B) optimization \cite{doi:10.1137/0916069} to quickly converge to a local optima around that region and refine the preliminary estimate obtained from LHS. This ensures a good estimate of the output bounds of the neural network.

Once these extrema estimates are obtained, they are fed into an SMT solver (Microsoft Z3 Theorem Prover \cite{10.1007/978-3-540-78800-3_24}) along with the safety violation properties to analyze the safety of the neural network.

\begin{itemize}
    \item If the analysis determines that a safety violation is not possible given the computed output bounds, the tool returns \texttt{holds}.
    \item If the analysis finds an LHS sample that is a valid safety violation, the tool returns \texttt{violated} along with the counterexample.
    \item If the tool encounters safety specifications with complex disjunctions on input variables, or if the analysis is inconclusive, the tool quits gracefully and returns \texttt{unknown}.
\end{itemize}

\texttt{BoxRL-NNV} also implements caching to reduce computational overheads over incremental runs.

\begin{algorithm}[h]
\caption{\texttt{BoxRL-NNV} Verification Algorithm}
\label{alg:boxrlnnv}
\begin{algorithmic}[1]
\REQUIRE Neural network $N$ (ONNX), safety specification $S$ (VNNLIB)
\ENSURE Verification result: \texttt{holds}, \texttt{violated} (with counterexample \texttt{CE}) or \texttt{unknown}

\STATE $inputSize \gets getInputSize(N)$
\STATE $Solver \gets Z3Solver(S)$
\IF{$complexDisjunction(Solver)$}
    \RETURN \texttt{unknown}
\ENDIF
\STATE $\mathcal{B} \gets extractInputBounds(Solver)$
\STATE $X \gets LHS\_MDU(20 \times inputSize, \mathcal{B})$
\STATE $R \gets extractOptima(N, X)$
\STATE $O \gets L\text{\_}BFGS\text{\_}B(R)$
\STATE $Solver.add(boundaryConstraints(O))$
\IF{$Solver.check() = unsat$}
    \RETURN \texttt{holds}
\ENDIF
\FOR{$x \in X$}
        \IF{$Solver.evaluate(x, N(x))$}
            \STATE $\texttt{CE} \gets Solver.model()$
            \RETURN \texttt{violated, CE}
        \ENDIF
\ENDFOR
\RETURN \texttt{unknown}
\end{algorithmic}
\end{algorithm}

\section{Results}

ACAS Xu is a family of 45 real-world deep neural networks (further grouped into 5 batches of 9 networks each), developed as an early prototype for the next-generation airborne collision avoidance system for unmanned aircraft. All of these networks are fully connected, having 8 layers and 300 ReLU nodes each \cite{10.1007/978-3-319-63387-9_5}. There are a total of 10 properties that can be verified for each network.

For the purposes of this paper, we evaluated \texttt{BoxRL-NNV} on one network from each batch for each of the 10 properties, that is, a total of 50 evaluations. The evaluation was done on a Google Colaboratory notebook with Python 3 Google Compute Engine backend (with CPU and 12.7GB RAM). \texttt{BoxRL-NNV} verified 20 instances as \texttt{holds} and 15 instances as \texttt{violated} (with \texttt{CE}), while returning \texttt{unknown} on 15 instances. The total time taken was 318.65 seconds.

\begin{table}[h]
\centering
\begin{tabular}{ |p{2.5cm}||p{2cm}|p{2.2cm}|  }
 \hline
 \multicolumn{3}{|c|}{\texttt{ACASXU\_run2a\_1\_1\_batch\_2000.onnx}} \\
 \hline
 \textbf{Safety Spec} & \textbf{Result} & \textbf{Time Taken (s)}\\
 \hline
 \texttt{prop\_1.vnnlib}   & \texttt{holds}    & 12.141\\
 \texttt{prop\_2.vnnlib}   & \texttt{holds}    & 0.953\\
 \texttt{prop\_3.vnnlib}   & \texttt{unknown}    & 7.265\\
 \texttt{prop\_4.vnnlib}   & \texttt{unknown}    & 10.028\\
 \texttt{prop\_5.vnnlib}   & \texttt{unknown}    & 10.955\\
 \texttt{prop\_6.vnnlib}   & \texttt{unknown}    & 1.042\\
 \texttt{prop\_7.vnnlib}   & \texttt{violated}    & 9.661\\
 \texttt{prop\_8.vnnlib}   & \texttt{violated}    & 6.801\\
 \texttt{prop\_9.vnnlib}   & \texttt{unknown}    & 7.888\\
 \texttt{prop\_10.vnnlib}   & \texttt{holds}    & 6.640\\
 \hline
\end{tabular}
\end{table}

\begin{table}[h]
\centering
\begin{tabular}{ |p{2.5cm}||p{2cm}|p{2.2cm}|  }
 \hline
 \multicolumn{3}{|c|}{\texttt{ACASXU\_run2a\_2\_1\_batch\_2000.onnx}} \\
 \hline
 \textbf{Safety Spec} & \textbf{Result} & \textbf{Time Taken (s)}\\
 \hline
 \texttt{prop\_1.vnnlib}   & \texttt{holds}    & 9.327\\
 \texttt{prop\_2.vnnlib}   & \texttt{violated}    & 0.762\\
 \texttt{prop\_3.vnnlib}   & \texttt{holds}    & 7.825\\
 \texttt{prop\_4.vnnlib}   & \texttt{holds}    & 7.093\\
 \texttt{prop\_5.vnnlib}   & \texttt{violated}    & 7.570\\
 \texttt{prop\_6.vnnlib}   & \texttt{unknown}    & 0.728\\
 \texttt{prop\_7.vnnlib}   & \texttt{violated}    & 7.402\\
 \texttt{prop\_8.vnnlib}   & \texttt{violated}    & 7.358\\
 \texttt{prop\_9.vnnlib}   & \texttt{unknown}    & 7.806\\
 \texttt{prop\_10.vnnlib}   & \texttt{holds}    & 7.118\\
 \hline
\end{tabular}
\end{table}

\begin{table}[h]
\centering
\begin{tabular}{ |p{2.5cm}||p{2cm}|p{2.2cm}|  }
 \hline
 \multicolumn{3}{|c|}{\texttt{ACASXU\_run2a\_3\_1\_batch\_2000.onnx}} \\
 \hline
 \textbf{Safety Spec} & \textbf{Result} & \textbf{Time Taken (s)}\\
 \hline
 \texttt{prop\_1.vnnlib}   & \texttt{holds}    & 7.794\\
 \texttt{prop\_2.vnnlib}   & \texttt{violated}    & 0.774\\
 \texttt{prop\_3.vnnlib}   & \texttt{holds}    & 7.624\\
 \texttt{prop\_4.vnnlib}   & \texttt{holds}    & 7.229\\
 \texttt{prop\_5.vnnlib}   & \texttt{unknown}    & 7.622\\
 \texttt{prop\_6.vnnlib}   & \texttt{unknown}    & 0.727\\
 \texttt{prop\_7.vnnlib}   & \texttt{violated}    & 6.769\\
 \texttt{prop\_8.vnnlib}   & \texttt{violated}    & 7.735\\
 \texttt{prop\_9.vnnlib}   & \texttt{violated}    & 7.774\\
 \texttt{prop\_10.vnnlib}   & \texttt{holds}    & 7.074\\
 \hline
\end{tabular}
\end{table}

\begin{table}[h]
\centering
\begin{tabular}{ |p{2.5cm}||p{2cm}|p{2.2cm}|  }
 \hline
 \multicolumn{3}{|c|}{\texttt{ACASXU\_run2a\_4\_1\_batch\_2000.onnx}} \\
 \hline
 \textbf{Safety Spec} & \textbf{Result} & \textbf{Time Taken (s)}\\
 \hline
 \texttt{prop\_1.vnnlib}   & \texttt{holds}    & 7.769\\
 \texttt{prop\_2.vnnlib}   & \texttt{holds}    & 0.752\\
 \texttt{prop\_3.vnnlib}   & \texttt{holds}    & 6.805\\
 \texttt{prop\_4.vnnlib}   & \texttt{holds}    & 7.594\\
 \texttt{prop\_5.vnnlib}   & \texttt{violated}    & 7.428\\
 \texttt{prop\_6.vnnlib}   & \texttt{unknown}    & 0.957\\
 \texttt{prop\_7.vnnlib}   & \texttt{violated}    & 6.882\\
 \texttt{prop\_8.vnnlib}   & \texttt{violated}    & 7.746\\
 \texttt{prop\_9.vnnlib}   & \texttt{unknown}    & 7.029\\
 \texttt{prop\_10.vnnlib}   & \texttt{holds}    & 7.611\\
 \hline
\end{tabular}
\end{table}

\begin{table}[h]
\centering
\begin{tabular}{ |p{2.5cm}||p{2cm}|p{2.2cm}|  }
 \hline
 \multicolumn{3}{|c|}{\texttt{ACASXU\_run2a\_5\_1\_batch\_2000.onnx}} \\
 \hline
 \textbf{Safety Spec} & \textbf{Result} & \textbf{Time Taken (s)}\\
 \hline
 \texttt{prop\_1.vnnlib}   & \texttt{holds}    & 7.655\\
 \texttt{prop\_2.vnnlib}   & \texttt{holds}    & 0.728\\
 \texttt{prop\_3.vnnlib}   & \texttt{unknown}    & 6.936\\
 \texttt{prop\_4.vnnlib}   & \texttt{holds}    & 7.796\\
 \texttt{prop\_5.vnnlib}   & \texttt{unknown}    & 7.664\\
 \texttt{prop\_6.vnnlib}   & \texttt{unknown}    & 0.966\\
 \texttt{prop\_7.vnnlib}   & \texttt{violated}    & 6.845\\
 \texttt{prop\_8.vnnlib}   & \texttt{violated}    & 7.690\\
 \texttt{prop\_9.vnnlib}   & \texttt{unknown}    & 6.817\\
 \texttt{prop\_10.vnnlib}   & \texttt{holds}    & 7.495\\
 \hline
\end{tabular}
\end{table}

\newpage
\section{Conclusion and Future Work}

For very large input dimensions, LHS can be computationally expensive in practice. Future work will explore approaches for improving scalability and handling inconclusive cases using surrogate modelling techniques to derive interpretable rules from neural network outputs.

\section*{Acknowledgments}
The author acknowledges Shubhajit Roy, Senior Research Fellow, IIT Gandhinagar and Avishek Lahiri, Senior Research Fellow, IACS Kolkata for their valuable input. The author also acknowledges Dr. Rajarshi Ray, Associate Professor, IACS Kolkata for his support and feedback.

\printbibliography
\end{document}